\documentclass[a4paper,twoside]{article}

\usepackage{epsfig}
\usepackage{subcaption}
\usepackage{calc}
\usepackage{amssymb}
\usepackage{amstext}
\usepackage{amsmath}
\usepackage{amsthm}
\usepackage{multicol}
\usepackage[nocompress]{cite}
\usepackage{pslatex}
\usepackage{apalike}
\usepackage{algorithm2e}
\usepackage[bottom]{footmisc}
\usepackage{mathptmx}
\usepackage[labelfont=bf]{caption}

\usepackage{SCITEPRESS}

\begin{document}

\title{Comparison of Text-Based and Image-Based Retrieval in Modern Multimodal Retrieval Augmented Generation Large Language Model Systems}

\author{
    \authorname{
        Elias Lumer,
        Alex Cardenas,
        Matt Melich, 
        Myles Mason,
        Sara Dieter, \\
        Vamse Kumar Subbiah,
        Pradeep Honaganahalli Basavaraju, 
        and Roberto Hernandez
    }
    \affiliation{\\PricewaterhouseCoopers U.S.}
}

\abstract{
Recent advancements in Retrieval-Augmented Generation (RAG) have enabled Large Language Models (LLMs) to access multimodal knowledge bases containing both text and visual information such as charts, diagrams, and tables in financial documents. However, existing multimodal RAG systems rely on LLM-based summarization to convert images into text during preprocessing, storing only text representations in vector databases, which causes loss of contextual information and visual details critical for downstream retrieval and question answering. To address this limitation, we present a comprehensive comparative analysis of two retrieval approaches for multimodal RAG systems, including text-based chunk retrieval (where images are summarized into text before embedding) and direct multimodal embedding retrieval (where images are stored natively in the vector space and passed to vLLMs during generation). We evaluate all three approaches across 6 LLM models and a two multi-modal embedding models on a newly created financial earnings call benchmark comprising 40 question-answer pairs, each paired with 2 documents (1 image and 1 text chunk). Experimental results demonstrate that direct multimodal embedding retrieval significantly outperforms LLM-summary-based approaches, achieving absolute improvements of 13\% in mean average precision (mAP@5) and 11\% in normalized discounted cumulative gain. These gains correspond to relative improvements of 32\% in mAP@5 and 20\% in nDCG@5, providing stronger evidence of their practical impact. We additionally find that direct multimodal retrieval and vLLM generation produce more accurate and factually consistent answers as measured by LLM-as-a-judge pairwise comparisons. We demonstrate that LLM summarization introduces information loss during preprocessing, whereas direct multimodal embeddings preserve visual context for retrieval and generation with vLLMs.
}

\keywords{Multimodal Retrieval-augmented generation, Large Language Models, Multimodal Embeddings, Vector Search}

\onecolumn \maketitle \normalsize \setcounter{footnote}{0} \vfill

\section{\uppercase{Introduction}}
\label{sec:introduction}

Recent advancements in Large Language Models (LLMs) have enabled powerful question answering systems that leverage external knowledge bases through Retrieval-Augmented Generation (RAG) \cite{lewis_retrieval-augmented_2020,gao_retrieval-augmented_2024,huang_retrieval-augmented_2024}. With RAG, these systems can retrieve relevant information from vector databases and inject that context into the model's prompt at inference time, improving factual accuracy and reducing hallucinations. Current RAG systems handle text documents effectively through dense retrieval methods \cite{karpukhin_dense_2020}, though they face significant challenges when applied to multimodal documents containing both text and visual information such as charts, diagrams, and tables in financial reports or presentations.

Despite advancements in RAG pipelines for text-based retrieval, a significant gap remains in effectively handling multimodal content \cite{mei_survey_2025,abootorabi_ask_2025}. Current multimodal RAG systems rely on LLM-based summarization to convert images into text during preprocessing, where a vision-language model generates textual descriptions of each image, and only these text summaries are stored in the vector database. This approach introduces information loss, as visual context, spatial relationships, and numerical precision are degraded or omitted during text conversion. Additionally, inference-time solutions such as direct multimodal embedding retrieval, where images are stored natively in the same vector space as text, remain underexplored in production RAG workflows.

Recent breakthroughs in multimodal embedding models such as CLIP \cite{radford_learning_2021} and Jina v4 \cite{gunther_jina-embeddings-v4_2025,jinaai_jina-embeddings-v4_2025} offer a promising alternative. These vision-language models can embed text and images into a unified semantic vector space, enabling text queries to retrieve both textual passages and visual content based on shared meaning. However, existing multimodal RAG research has not systematically compared direct image embedding retrieval with conventional LLM-summary-based approaches across full end-to-end workflows encompassing retrieval accuracy, answer quality, and model robustness, particularly for financial documents \cite{gong_mhier-rag_2025,gondhalekar_multifinrag_2025,setty_improving_2024}.

In this paper, we present a comprehensive empirical comparison of multimodal retrieval strategies for RAG systems, evaluating text-based chunk retrieval (LLM-summary-based) and direct multimodal embedding retrieval. We evaluate both approaches across 6 LLM models and 2 embedding models on a newly created financial earnings call benchmark comprising 40 question-answer pairs, each paired with targeted documents (answer-relevant images and text chunks). Our evaluation spans both retrieval performance (Precision@5, Recall@5, mean Average Precision (mAP@5), normalized Discounted Cumulative Gain (nDCG@5)) and end-to-end answer quality through LLM-as-a-judge pairwise comparisons across six criteria: correctness, numerical fidelity, missing information, unsupported additions, conciseness, and clarity.

Experimental results demonstrate that direct multimodal embedding retrieval significantly outperforms text-based approaches, achieving 13\% absolute improvement in mean average precision and 11\% improvement in normalized discounted cumulative gain. These correspond to relative improvements of 32\% in mAP@5 and approximately 20\% in nDCG@5, indicating substantial ranking and relevance gains from preserving visual information in native form. When retrieved context is provided to downstream models, image-based retrieval leads to more accurate and factually consistent answers, particularly for larger models with stronger multimodal reasoning capabilities. These findings suggest that preserving visual information in its native form substantially improves both retrieval precision and generated response quality within multimodal RAG systems.

\section{\uppercase{Related Works}}
\label{sec:relatedwork}

\subsection{Retrieval-Augmented Generation}

Retrieval-Augmented Generation (RAG) has emerged as an effective approach to improve factual accuracy and reduce hallucinations in large language models by grounding generation in retrieved external knowledge \cite{lewis_retrieval-augmented_2020}. Lewis et al. introduced the foundational RAG architecture combining a dense retriever with a generative model for knowledge-intensive question answering, demonstrating that retrieval strengthens output grounding \cite{lewis_retrieval-augmented_2020}. Building on this foundation, dense passage retrieval methods such as DPR leverage bi-encoders to map queries and documents into shared embedding spaces, enabling semantic matching beyond lexical overlap \cite{karpukhin_dense_2020}. Recent surveys comprehensively review RAG techniques and confirm that while text-based RAG improves factual recall, most systems remain limited to textual content, leaving significant gaps in handling structured, spatial, or visual information common in complex documents \cite{gao_retrieval-augmented_2024,huang_retrieval-augmented_2024}. Advanced RAG methods further enhance retrieval through query rewriting \cite{ma_query_2023}, hypothetical document embeddings \cite{gao_precise_2022}, corrective retrieval \cite{yan_corrective_2024}, self-reflective generation \cite{asai_self-rag_2023}, and hybrid retrieval strategies combining semantic and lexical approaches \cite{sawarkar_blended_2024}. Reranking techniques using LLMs as re-ranking agents improve retrieval precision by reordering candidates based on query-document relevance \cite{sun_is_2023}. However, these advances predominantly focus on text-based documents, motivating extensions to multimodal content where visual and textual information must be jointly indexed and retrieved.

\subsection{Multimodal Embedding Models}

Multimodal embedding models enable unified representations of text and images in shared semantic spaces, facilitating cross-modal retrieval. CLIP pioneered vision-language pretraining by learning transferable visual representations from natural language supervision, aligning image and text embeddings through contrastive learning on large-scale web data \cite{radford_learning_2021}. Recent advances extend multimodal embeddings to support multilingual and task-specific retrieval. Jina Embeddings v3 introduced task-specific Low-Rank Adaptation (LoRA) for text embeddings, enabling flexible adaptation across diverse retrieval scenarios \cite{sturua_jina-embeddings-v3_2024}. Jina Embeddings v4 further advances this capability with universal embeddings supporting multimodal and multilingual retrieval, enabling direct comparison of text queries against image content in a unified vector space \cite{gunther_jina-embeddings-v4_2025}. These models eliminate the need for separate text and image encoders, providing a principled approach to multimodal retrieval without intermediate text conversion. Despite these architectural advances, systematic empirical comparisons evaluating how direct multimodal embeddings perform against conventional text-based image summarization in end-to-end RAG workflows remain limited.

\subsection{Document Understanding and Visual Preprocessing}

Several approaches address multimodal document understanding by converting visual content into text representations. Donut introduced an OCR-free document understanding transformer that learns from visual tokens rather than recognized text, reducing transcription errors but ultimately producing text sequences that discard visual layout and spatial relationships \cite{kim_ocr-free_2021}. Pix2Struct further advanced chart and interface understanding through screenshot parsing as pretraining, demonstrating improved visual reasoning with language supervision \cite{lee_pix2struct_2022}. While these methods make visual content searchable through text, they inherently remove layout, scale, and numeric precision critical for financial documents and data-rich visualizations \cite{setty_improving_2024}. Industry frameworks from Microsoft Azure emphasize multimodal RAG for complex document structures, advocating for Document Intelligence to extract and structure visual content before indexing \cite{microsoft_build_2024,microsoft_docintelligence_2024}. Practitioner guides similarly recommend multimodal pipelines for production systems \cite{analyticsvidhya_multimodal_2024,analyticsvidhya_rag_azure_2024,simplai_multimodal-production-rag_2024}. However, these approaches predominantly rely on preprocessing images into text summaries, introducing potential information loss that our work systematically evaluates.

\subsection{Multimodal RAG Systems}

Recent research has begun integrating text and visual modalities within RAG frameworks for document question answering. Comprehensive surveys on multimodal RAG highlight the growing interest in systems that handle diverse modalities including text, images, tables, and charts \cite{abootorabi_ask_2025,mei_survey_2025,multimodalrag_survey_site_2025}. MHier-RAG proposed hierarchical and multi-granularity reasoning for visual-rich document question answering, indexing both text and image data to improve document-level retrieval \cite{gong_mhier-rag_2025}. However, its image features derive from text captions and descriptions rather than native image embeddings, potentially losing visual fidelity. MultiFinRAG introduced an optimized multimodal RAG framework for financial question answering where figures and tables are summarized into structured text before indexing, demonstrating improvements on financial datasets \cite{gondhalekar_multifinrag_2025}. Similarly, work on improving retrieval for financial documents emphasizes preprocessing strategies to extract textual representations from complex layouts \cite{setty_improving_2024}. While these approaches demonstrate the value of incorporating visual context, they uniformly depend on LLM-generated summaries or OCR-extracted text as intermediaries, raising questions about information preservation and retrieval fidelity. Our work directly addresses this gap by empirically comparing text-based retrieval against direct multimodal embedding retrieval across retrieval metrics and downstream answer quality, isolating the impact of embedding modality choice within a controlled experimental framework.

\subsection{Evaluation Methodologies for Retrieval Systems}

Evaluation of retrieval systems relies on established information retrieval metrics to measure ranking quality and relevance. Normalized Discounted Cumulative Gain (nDCG) has become a standard metric for evaluating ranked retrieval results, accounting for both relevance and position in the ranking \cite{jarvelin_cumulated_2002}. Traditional lexical retrieval baselines including TF-IDF \cite{papineni_why_2001} and BM25 \cite{robertson_probabilistic_2009} provide reference points for dense retrieval evaluation. For end-to-end system evaluation, recent work introduces LLM-as-a-judge methodologies where language models assess response quality through pairwise comparisons, enabling scalable evaluation beyond exact string matching \cite{zheng_judging_2023}. Our evaluation framework combines standard retrieval metrics (Precision@5, Recall@5, mAP@5, nDCG@5) with LLM-as-a-judge assessment across correctness, numerical fidelity, completeness, conciseness, and clarity to provide comprehensive evaluation spanning retrieval accuracy and downstream generation quality.

\section{\uppercase{Methods}}
\label{sec:methods}

In this section, we present our comparative evaluation framework for multimodal retrieval strategies in RAG systems. Our methodology consists of three main components: (1) a manually curated financial earnings benchmark with multimodal ground truth annotations (\ref{subsec:dataset}), (2) two retrieval approaches spanning LLM-summary-based and direct multimodal embedding strategies (\ref{subsec:approaches}), and (3) a comprehensive evaluation framework combining retrieval metrics and LLM-as-a-judge answer quality assessment (\ref{subsec:evaluation}).

\subsection{Dataset Construction}\label{subsec:dataset}

We construct a financial earnings benchmark consisting of 40 multimodal question-answer pairs targeting information from a Fortune 500 company's publicly available earnings calls. Unlike existing financial QA datasets that focus solely on text, our benchmark explicitly requires integration of both textual and visual (graph) information to answer questions correctly.

\subsubsection{Data Collection}

Financial documents were sourced from quarterly earnings calls of a Fortune 500 company, including both earnings call transcripts (text documents) and corresponding investor presentation slide decks (visual documents). These materials contain critical financial information presented across modalities: narrative explanations in transcripts and data visualizations such as revenue charts, margin breakdowns, and growth metrics in slide decks. Each document collection represents a complete earnings event, ensuring that paired text and visual content share temporal and topical coherence.

\subsubsection{Question-Answer Pair Generation}

Forty multimodal question-answer pairs were manually created to reflect realistic financial analyst queries requiring multi-hop reasoning across both text and images. Questions were designed to necessitate retrieval and synthesis of information from both earnings transcripts and presentation slides. For each question, ground truth answers were manually annotated along with relevant page numbers from both document types, establishing explicit retrieval targets for evaluation. Each question is paired with their relevant text chunks and relevant images, corresponding to specific pages in the transcript and slide deck respectively. This paired structure enables direct comparison of text-only and multimodal retrieval performance on identical information needs.

\subsubsection{Document Preprocessing}

Earnings call transcripts were segmented into passages corresponding to logical sections or speaker turns. Presentation slide decks were converted to individual slide images, with each slide treated as a distinct retrievable unit. This preprocessing ensures that both modalities are indexed at comparable granularity, where each retrievable unit represents a coherent information block. Ground truth relevance labels for retrieval evaluation were assigned based on page numbers, enabling automated computation of precision, recall, and ranking metrics.

\subsection{Two Retrieval Approaches}\label{subsec:approaches}

We compare two retrieval strategies representing different approaches to handling multimodal content in RAG systems. Both approaches utilize Azure AI Search as the vector database backend and retrieve the top-5 most relevant documents for each query.
 
\subsubsection{Approach 1: Text-only Retrieval}

In this approach, visual content from the earnings presentation is converted into text before retrieval. Each slide image is passed to a OpenAI GPT-5 model to produce a textual description intended to capture the key information present in the visual, including chart labels, numerical values, and high-level context. These text descriptions serve as surrogates for the original images, a common strategy in production RAG systems \cite{microsoft_build_2024,analyticsvidhya_multimodal_2024}

Both the earnings call transcript chunks and the LLM slide descriptions are embedded using OpenAI text-embedding-ada-002, a widely adopted dense embedding model for semantic retrieval \cite{lumer2025scalemcpdynamicautosynchronizingmodel,lumer2025graphragtoolfusion,lumer2025memtooloptimizingshorttermmemory,chen2024reinvoketoolinvocationrewriting,huang2024planningeditingretrieveenhanced,lumer2024toolshedscaletoolequippedagents,lumer2025tool}. At query time, user queries are also embedded using the same embedding model and matched against this text-only representation of the full document set. Because visual information is represented indirectly through generated text rather than native image embeddings, this approach may omit spatial structure, layout, or numeric precision present in the original image. 

This configuration reflects the current typical multimodal RAG practice of converting images into text before retrieval, and therefore serves as the comparison point for evaluating the benefits of direct image embedding in our experiment.

\subsubsection{Approach 2: Direct Multimodal Embedding Retrieval}

The direct multimodal embedding approach leverages Jina Embeddings v4 \cite{gunther_jina-embeddings-v4_2025}, a unified multimodal embedding model that maps both text and images into a shared semantic vector space. Unlike the text-based approach, images are stored natively in their visual form without intermediate text conversion. Both earnings call transcript chunks and slide deck images are embedded directly using Jina v4 and indexed in the same Azure AI Search vector database. At query time, text queries are embedded into the same multimodal space, enabling semantic retrieval of both textual passages and visual content based on shared meaning. This approach preserves visual information in its native representation, avoiding information loss from text conversion while enabling cross-modal retrieval. Images retrieved by this method are provided directly to downstream vision-language models during answer generation, allowing models to interpret visual content with full fidelity.

\subsection{Evaluation Framework}\label{subsec:evaluation}

Our evaluation spans both retrieval performance and end-to-end answer quality, providing comprehensive assessment of how embedding modality choice impacts multimodal RAG systems.

\subsubsection{Models Evaluated}

We evaluate six OpenAI language models spanning multiple capability tiers: OpenAI GPT-4o, GPT-4o-mini, GPT-4.1, GPT-4.1-mini, GPT-5, and GPT-5-mini. These models represent varying levels of reasoning capability and multimodal understanding, enabling analysis of how model scale interacts with retrieval strategy effectiveness. For both approaches, the embedding models remains constant while the downstream LLM varies. This design isolates the impact of retrieval modality while controlling for embedding model choice, reflecting prior findings that embedding model variation has negligible impact on retrieval performance in text-only settings \cite{lumer2025scalemcpdynamicautosynchronizingmodel,chen2024reinvoketoolinvocationrewriting,huang2024planningeditingretrieveenhanced,lumer2025graphragtoolfusion,lumer2025memtooloptimizingshorttermmemory}.

\subsubsection{Retrieval Metrics}

Retrieval performance is measured using four standard information retrieval metrics computed over the top-5 retrieved documents. Precision@5 measures the fraction of retrieved documents that are relevant, while Recall@5 measures the fraction of all relevant documents successfully retrieved. Mean Average Precision (mAP@5) computes the average precision across all queries, accounting for ranking order. Normalized Discounted Cumulative Gain (nDCG@5) evaluates ranking quality by assigning higher weight to relevant documents appearing earlier in the ranking \cite{jarvelin_cumulated_2002}. Ground truth relevance is determined by page numbers annotated during dataset construction: a retrieved document is considered relevant if its page number matches the ground truth page number for the corresponding question. These metrics provide a comprehensive view of retrieval accuracy, coverage, and ranking quality across the two approaches.

\subsubsection{Answer Quality Assessment}

End-to-end answer quality is evaluated using LLM-as-a-judge methodology \cite{zheng_judging_2023}, where OpenAI GPT-5 performs pairwise comparisons between answers generated by the text-based approach and the direct multimodal embedding approach. For each of the 40 questions, both approaches retrieve relevant context and generate answers using the same downstream LLM. OpenAI GPT-5 then evaluates answer pairs across six binary criteria: Correctness (factual alignment with ground truth), Numerical Fidelity (accuracy of numeric values), Missing Information (content completeness), No Unsupported Additions (absence of hallucinations), Conciseness (efficient wording), and Clarity (readability). For each criterion, the judge assigns a score of 1 to the preferred answer and 0 to the other, enabling aggregation across questions and models. This evaluation design isolates the impact of retrieval modality on downstream generation quality when both approaches retrieve multimodal content but differ in how images are represented during retrieval.

\subsubsection{Answer Generation Pipeline}

Retrieved context from each approach is provided to downstream language models through a standard RAG prompt template. The prompt includes the user question, retrieved text chunks and image summaries (for text-only approach), or retrieved text chunks and native images (for multimodal approach), and instructs the model to generate an answer grounded in the provided context. For the multimodal approach, image quality is handled by the OpenAI API based on highest allowed resolution settings, following standard vision model preprocessing \cite{openai_images_vision_detail_2024}. This pipeline ensures that differences in answer quality arise from retrieval strategy rather than prompt engineering or generation parameters, providing a controlled comparison of how embedding modality impacts end-to-end RAG performance.

\begin{figure*}[t]
\centering
\fbox{\includegraphics[width=0.8\textwidth]{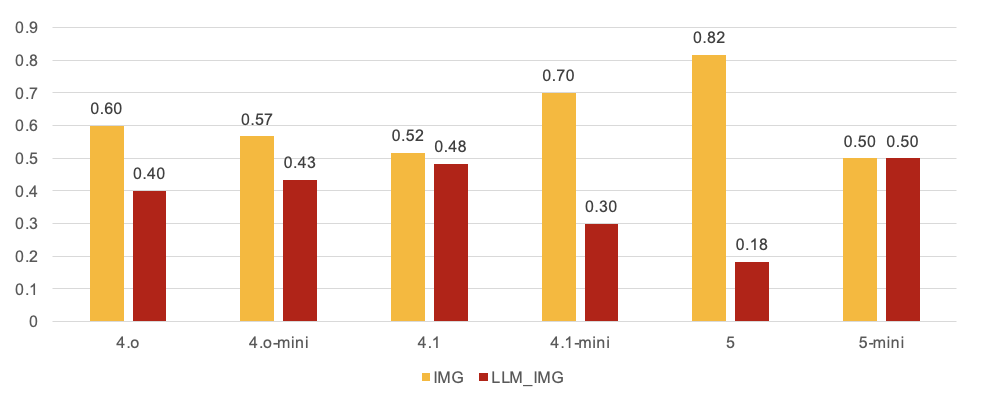}}
\caption{Averaged pairwise comparison scores across all six LLM models. Yellow bars represent IMG (direct multimodal embedding retrieval) and red bars represent LLM\_IMG (LLM-summary-based retrieval). Scores indicate win rate proportions, with IMG consistently outperforming LLM\_IMG across all models, particularly for larger non-mini variants.}
\label{fig:judge_scores_all}
\end{figure*}

\section{\uppercase{Experiments}}
\label{sec:experiments}

\subsection{Experimental Settings}

We evaluate the two retrieval approaches on our financial earnings benchmark consisting of 40 multimodal question-answer pairs, each associated with relevant text chunks and relevant images. Retrieval performance is measured using four standard information retrieval metrics computed over the top-5 retrieved documents: Precision@5, Recall@5, mAP@5, and nDCG). For end-to-end answer quality evaluation, we generate answers using six OpenAI language models: GPT-4o, GPT-4o-mini, GPT-4.1, GPT-4.1-mini, GPT-5, and GPT-5-mini. OpenAI GPT-5 serves as the judge model for pairwise comparisons, evaluating answers across six binary criteria: Correctness (factual alignment), Numerical Fidelity (number accuracy), Missing Information (content completeness), No Unsupported Additions (hallucination control), Conciseness (efficient wording), and Clarity (readability). Each criterion receives a score of 1 for the preferred answer and 0 for the other. All experiments use Azure AI Search as the vector database backend with OpenAI text-embedding-ada-002 for text-only approaches and Jina Embeddings v4 for multimodal retrieval.

\subsection{Retrieval Performance Results}

Table \ref{tab:retrieval_results} presents the macro-averaged retrieval performance comparing direct multimodal embedding retrieval (IMG) against text-based image retrieval (LLM\_IMG). Direct multimodal retrieval significantly outperforms the text-LLM-summary approach across all metrics. The multimodal approach achieves mAP@5 of 0.5234 compared to 0.3963 for the text-based approach, representing an improvement of 0.1271 or 32\% relative gain. Similarly, the multimodal approach obtains nDCG@5 of 0.6543 compared to 0.5448 for the text-based approach, showing an improvement of 0.1095 or 20\% relative gain. Precision@5 improves from 0.480 to 0.540, an increase of 0.060 or 12.5\%, while Recall@5 increases from 0.5362 to 0.5529, an improvement of 0.0167 or 3\%. These results demonstrate that direct image embeddings capture relevant documents more effectively than purely text-based embeddings. The highest gains appear in mAP@5 and nDCG@5, indicating that multimodal embeddings not only retrieve more relevant documents but also produce superior ranking quality, placing the most relevant documents higher in the result list. This improved ranking directly benefits downstream answer generation by providing models with better-ordered context.

\begin{table}[ht]
\centering
\resizebox{0.48\textwidth}{!}{%
\begin{tabular}{lcccc}
\hline
\textbf{Method} & \textbf{Precision@5} & \textbf{Recall@5} & \textbf{mAP@5} & \textbf{nDCG@5} \\
\hline
IMG & \textbf{0.540} & \textbf{0.5529} & \textbf{0.5234} & \textbf{0.6543} \\
LLM\_IMG & 0.480 & 0.5362 & 0.3963 & 0.5448 \\
\hline
\end{tabular}
} % end resizebox
\caption{Comparison of macro-averaged retrieval results for direct multimodal embedding retrieval (IMG) and text-based image retrieval (LLM\_IMG). Bold indicates best performance. Direct multimodal embeddings achieve substantial improvements in mAP@5 and nDCG@5, demonstrating superior ranking quality.}
\label{tab:retrieval_results}
\end{table}

\subsection{Answer Quality Results}

End-to-end answer quality was evaluated through pairwise comparisons between the multimodal and text-based approaches using OpenAI GPT-5 as the judge model. Table \ref{tab:judge_scores} presents the averaged judge scores across all six LLM models, where scores represent the proportion of times each approach was preferred. Figure \ref{fig:judge_scores_all} visualizes these averaged results, showing consistent preference for the multimodal approach across all models.

The multimodal approach achieves an overall average win rate of 0.612 compared to 0.388 for the LLM-summary approach when averaged across all criteria and models. The gap is particularly pronounced for larger models: GPT-5 shows the strongest preference for the multimodal approach with an average score of 0.82 compared to 0.18 for the LLM-summary approach, while OpenAI GPT-4o achieves 0.60 compared to 0.40 and o1 obtains 0.52 compared to 0.48. In contrast, smaller mini models exhibit more balanced scores, with OpenAI GPT-4o-mini at 0.57 compared to 0.43, o1-mini at 0.70 compared to 0.30, and o5-mini at 0.50 compared to 0.50, indicating that these models derive limited benefit from enhanced multimodal context during answer generation.

\begin{table}[ht]
\centering
\small
\caption{Averaged pairwise comparison scores across all six LLM models. Scores represent the proportion of times each approach was preferred by OpenAI GPT-5 across six evaluation criteria. Higher scores indicate stronger preference.}
\label{tab:judge_scores}
\begin{tabular}{lcc}
\hline
\textbf{Model} & \textbf{IMG} & \textbf{LLM\_IMG} \\
\hline
OpenAI GPT-4o & 0.60 & 0.40 \\
OpenAI GPT-4o-mini & 0.57 & 0.43 \\
OpenAI GPT-4.1 & 0.52 & 0.48 \\
OpenAI GPT-4.1-mini & 0.70 & 0.30 \\
OpenAI GPT-5 & 0.82 & 0.18 \\
OpenAI GPT-5-mini & 0.50 & 0.50 \\
\hline
\textbf{Average} & \textbf{0.612} & \textbf{0.388} \\
\hline
\end{tabular}
\end{table}

\begin{figure*}[t]
\centering
\fbox{\includegraphics[width=0.78\textwidth]{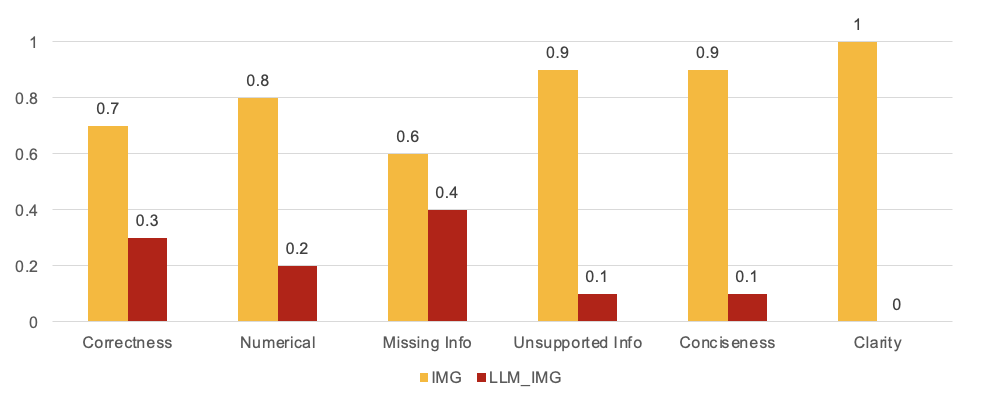}}
\caption{Breakdown of pairwise comparison scores for GPT-5 across six evaluation criteria. Yellow bars represent IMG and red bars represent LLM\_IMG. IMG shows substantial advantages in Correctness, Numerical Fidelity, and No Unsupported Additions (hallucination control), demonstrating that native image embeddings preserve critical information lost during text conversion.}
\label{fig:judge_scores_gpt5}
\end{figure*}

Figure \ref{fig:judge_scores_gpt5} breaks down the pairwise comparison results for GPT-5 across the six evaluation criteria. The multimodal approach demonstrates substantial advantages in Correctness (0.70 compared to 0.30), Numerical Fidelity (0.80 compared to 0.20), and No Unsupported Additions (0.90 compared to 0.10), indicating that direct multimodal retrieval produces more factually accurate answers with fewer hallucinations.

Missing Information scores favor the multimodal approach (0.60 compared to 0.40), suggesting more complete answers when visual context is preserved. Conciseness shows a near tie (0.90 compared to 0.10), while Clarity achieves perfect preference for the multimodal approach (1.00 compared to 0.00). The most striking result appears in hallucination control, where the multimodal approach prevents unsupported additions 90\% of the time, substantially reducing the information loss and fabrication that occurs when images are converted to text summaries during preprocessing.

\subsection{Discussion}

The experimental results demonstrate that direct multimodal embedding retrieval substantially outperforms text-based approaches across both retrieval metrics and downstream answer quality. The 32\% relative improvement in mAP@5 and 20\% improvement in nDCG@5 indicate that preserving visual information in its native form enables more accurate semantic matching between queries and multimodal documents. This improved retrieval accuracy translates directly to higher-quality generated answers, particularly for larger models with stronger multimodal reasoning capabilities. The pronounced gap in hallucination control, where the multimodal approach reduces unsupported additions by 80 absolute percentage points for GPT-5, suggests that LLM summarization introduces not only information loss but also fabricated details that propagate to downstream generation. The diminishing returns observed for mini models indicate that effective utilization of multimodal context requires sufficient model capacity for cross-modal reasoning. These findings have practical implications for production RAG systems handling multimodal documents: while LLM summarization offers a convenient preprocessing strategy compatible with existing text-only infrastructure, it fundamentally limits retrieval and generation quality compared to native multimodal embeddings. Organizations should prioritize multimodal embedding models when deploying RAG systems for document types where visual information carries critical semantics, particularly in domains such as financial reporting where charts, tables, and numerical visualizations convey information difficult to capture through text alone.

\section{\uppercase{Limitations}}
\label{sec:limitations}

While direct multimodal embedding retrieval advances multimodal RAG systems, a key limitation concerns preprocessing complexity for multimodal embeddings. Unlike text-based approaches that convert images to text through a single LLM call, multimodal embedding pipelines require explicit image detection, extraction, and format conversion steps. Documents must be parsed to identify charts, tables, and images, with each visual element saved as a separate file before embedding. This preprocessing burden increases for diverse document types, as PowerPoint presentations where entire slides serve as retrievable units differ fundamentally from PDF reports where individual figures must be extracted. Automated preprocessing tools such as Docling \cite{docling_document_processing_2024}, Azure Document Intelligence \cite{azure_document_intelligence_2024}, and Unstructured.io \cite{unstructured_io_data_ingestion_2024} provide partial solutions, but distinguishing between tables and images remains challenging. Future work should develop robust document parsing pipelines that automatically segment and classify visual elements across document formats, reducing the operational overhead of deploying multimodal RAG systems in production environments.

\section{\uppercase{Conclusion}}
\label{sec:conclusion}

Multimodal RAG systems must effectively retrieve and reason over both textual and visual content from documents containing charts, tables, and images. We present a comparative evaluation of two retrieval approaches for multimodal RAG systems, including text-based chunk retrieval where images are converted to text during preprocessing, and direct multimodal embedding retrieval where images are stored natively in vector space. We evaluated both approaches across six OpenAI language models on a newly created financial earnings benchmark comprising 40 question-answer pairs requiring integration of textual and visual information. Experimental results demonstrate that direct multimodal embedding retrieval substantially outperforms text-based approaches, achieving a 32\% relative improvement in mean average precision and an overall win rate of 0.612 compared to 0.388 in LLM-as-a-judge pairwise comparisons. These findings provide empirical evidence that preserving visual information in native form rather than converting to text summaries enables more accurate retrieval and downstream generation in multimodal RAG systems. Future work should extend evaluation to diverse domains including medical, legal, and scientific documents where visual content serves different purposes, while developing automated pipelines to reduce operational overhead. As multimodal embedding models mature and vision-language models strengthen their cross-modal reasoning capabilities, the performance advantages of direct multimodal retrieval are likely to widen further.

\bibliographystyle{apalike}
{\small
\bibliography{references}}

\end{document}